# Optimizing PM2.5 Forecasting Accuracy with Hybrid Meta-Heuristic and Machine Learning Models

Parviz Ghafariasl* , Masoomeh Zeinalnezhad[∥], *Amir Ahmadishokooh*[V]

*Department of Industrial and Manufacturing Systems Engineering*

*Kansas State University, Manhattan, KS, 66506, USA.*

*\*Parvizghafari@ksu.edu*

[∥] *Department of Industrial Engineering, West Tehran Branch,*

*Islamic Azad University, Tehran,Iran.*

*zeinalnezhad.m@wtiau.ac.ir*

[V]*Department of Industrial Engineering, Science and Research Branch, Islamic Azad University,*

*Tehran, Iran.*

*amir.ahmadi.shokooh@gmail.com*

**Abstract:** Timely alerts about hazardous air pollutants are crucial for public health. However, existing forecasting models often overlook key factors like baseline parameters and missing data, limiting their accuracy. This study introduces a hybrid approach to address these issues, focusing on forecasting hourly PM2.5 concentrations using Support Vector Regression (SVR). Meta-heuristic algorithms, Grey Wolf Optimization (GWO) and Particle Swarm Optimization (PSO), optimize SVR Hyper-parameters "C" and "Gamma" to enhance prediction accuracy. Evaluation metrics include R-squared (R2), Root Mean Square Error (RMSE), and Mean Absolute Error (MAE). Results show significant improvements with PSO-SVR (R2: 0.9401, RMSE: 0.2390, MAE: 0.1368) and GWO-SVR (R2: 0.9408, RMSE: 0.2376, MAE: 0.1373), indicating robust and accurate models suitable for similar research applications.

## 1. Introduction

Historically, when the amount of produced data was limited, many managers and decision-makers understood the concepts behind them with a superficial look and manual separation of the data (Zeinalnezhad et al., 2019). Due to the great importance of data and the progress of data mining, this study has encountered the production of a large amount of them. The science of data mining has provided a platform that can be used to classify, analyse and extract the hidden concepts in the data by using new technologies, such as artificial intelligence and machine learning, appropriate for specific goals and to use them to make critical decisions (Goodarzi et al., 2022). Various businesses around the world generate enormous data sets (Chofreh et al., 2021). These data sets include sales transactions, marketing data, product information, advertisements, company records and reports, and customer feedback (Alavi et al., 2022). In meteorology and air pollution, relevant organisations and researchers have been trying to identify and produce appropriate data to predict the intensity of pollutants and protect human health from them (Kong et al. 2024).

The increase in the earth's temperature, climate change, and the rise of the sea level are among the consequences of the high consumption of fossil energy and the release of pollutants and greenhouse gases (Lopes et al. 2024). The depletion of fossil energy resources and the projected rise in prices have prompted legislators to devise regulations and policies for environmental management. Simultaneously, academics are actively pursuing the development of cleaner and renewable energy sources as potential alternatives to the existing energy infrastructure. The prediction of carbon dioxide ($CO_2$) emissions has the utmost importance due to the substantial impact it has on climate change and global warming, with the grave hazards it poses to human health (Shabani et al. 2021).

The provision of air quality forecasting serves as an efficient means of safeguarding public health by offering timely alerts regarding the presence of detrimental air pollutants (Bai et al. 2018). The ongoing concern of governments and individuals is focused on the degradation of air quality, the frequent occurrence of air contaminants, and the consequent health impacts (Zeinalnezhad et al. 2021). There is a pressing need for the development of suitable and efficient forecasting tools within the realm of scientific study. Forecasting models are subject to continuous improvement and expansion. Various methodologies and approaches can be explored in the context of time series data analysis to effectively address air pollution concerns and mitigate the occurrence of severe pollution episodes (Liu et al. 2021). Various algorithms and methods, such as time series models, machine learning algorithms, neural networks, deep learning, and heuristic and meta-heuristic algorithms, have been used to predict pollutants (Zeinalnezhad et al. 2020) (Lai et al. 2023). According to these contents, this study aims to introduce a better possible algorithm with a minor error to predict the emission of pollutants and obtain results with wider dimensions and the closest to the real world.

The correlation between energy usage and the release of pollutants and greenhouse gases is widely acknowledged. The utilisation of fossil fuels, encompassing petroleum, petroleum byproducts, and natural gas, results in the emission and augmentation of greenhouse gases, namely carbon dioxide (Ghafariasl et al. 2024). Developing nations significantly contribute to the production and dissemination of greenhouse gas emissions. China, the United States of America, Russia, India, Japan, Germany, Canada, Great Britain, South Korea, and Iran are internationally acknowledged as the foremost 10 global participants in carbon dioxide emissions (Qiao et al. 2021).

Due to the severe dangers of air pollution to human health and natural ecosystems, various studies have predicted these pollutants as an essential policy tool. Boznar et al. (Boznar, Lesjak, and Mlakar 1993) were the first to model hourly sulfur dioxide ($SO_2$) concentration in polluted industrial areas of Slovenia with a neural network. Their study used the input parameters of temperature, wind speed, direction, solar radiation, and time of day. Ultimately, the results showed the usefulness of using neural networks in modeling. Juhos et al. (Juhos, Makra, and Tóth 2008) predicted NO and $NO_2$ concentration values for the next four days using MLP neural networks and

SVR along with the PCA preprocessing method and showed that both SVR and MLP neural network models can predict NO and $NO_2$ pollutants.

In their study, Pao and Tsai (2011) conducted an analysis on the interconnections among pollutant emissions, energy consumption, and output in Brazil from 1980 to 2007. The gray forecasting model (GM) forecasts three variables during 2008-2013. The non-linear gray Bernoulli model (NGBM) predicts three indicators of carbon dioxide emissions, energy consumption, and actual outputs (Pao, Fu, and Tseng 2012). A numerical iterative approach is proposed for the optimisation of the NGBM parameter. The present study examines the "full breakdown" technique to assess the intensity of CO2 emissions and its constituent components across 36 economic sectors throughout the period spanning from 1996 to 2009 (Robaina Alves and Moutinho 2013). A novel accounting methodology was employed, using forecast error variance decomposition and shock response functions to analyse the elements that contribute to the decomposition of emission intensity. Yang and Zhao (2014) conducted a study whereby they analysed the temporal correlations between economic growth, energy consumption, and carbon emissions in India from 1970 to 2008. The researchers utilised sophisticated methodologies, including out-of-sample Granger causality tests and directed acyclic graphs (DAG), to examine these relationships.

In an independent investigation, Wu et al. (2015) undertook an examination of the interrelationship between energy consumption, urban population, economic growth, and $CO_2$ emissions among the BRICS nations (Brazil, Russia, India, China, and South Africa) throughout the period spanning from 2004 to 2010. The researchers employed a New Multivariate Grey Revisited model for their investigation. This study aims to examine the various impacts via which carbon dioxide ($CO_2$) emissions in the tourist sector can be analysed. Specifically, it seeks to investigate the evolution of these effects over time and determine which of them play a more significant role in determining the overall emissions. (Robaina-Alves, Moutinho, and Costa 2016). In this study, the analysis technique employed was the logarithmic average division index, which was applied to examine five distinct sub-sectors within the tourism industry in Portugal during the period from 2000 to 2008. In their study, Wang and Ye (2017) proposed a multivariate grey model that incorporates the power exponential expression of key factors as exogenous variables to forecast carbon dioxide emissions resulting from fossil energy consumption. Two non-linear programming models are formulated with the objective of minimising the average absolute percentage of error. The purpose of these models is to determine the values of the unknown parameters in the non-linear grey multivariate model. Moreover, for the purpose of improving the suitability of the Grey model for datasets with large sample sizes, the data about China's Gross Domestic Product (GDP) and carbon emissions arising from the consumption of fossil energy between the years 1953 and 2013 is divided into fifteen separate stages.

In their study, Fand et al. (2018) introduced an enhanced Gaussian process regression technique for predicting carbon dioxide emissions. This approach incorporates a modified particle swarm optimisation (PSO) algorithm to efficiently optimise the parameters of the covariance function in Gaussian process regression. The authors also conducted experiments using their enhanced Particle Swarm Optimization-Gaussian Process Regression (PSO-GPR) technique using comprehensive data pertaining to total carbon dioxide ($CO_2$) emissions in the United States, China, and Japan for the period spanning from 1980 to 2012. The authors conducted a comparative analysis of the predictive performance of their proposed methodology in relation to Gaussian Process Regression (GPR) and the original Backpropagation (BP) neural networks. The evaluation was carried out using datasets sourced from the United States, China, and Japan. Hosseini et al. (2019) utilised Multiple Linear Regression (MLR) and Polynomial Regression (MPR) methodologies to predict the levels of carbon dioxide ($CO_2$) emissions in Iran for the year 2030. The researchers examined two specific scenarios, referred to as Business As Usual (BAU) and the Sixth Development Plan (SDP). In their study, Acheampong and Boateng (2019) utilised an Artificial Neural Network (ANN) approach to develop predictive models for carbon emission intensity in five specific countries, namely Australia, Brazil, China, India, and the United States of America. The researchers utilised a collection of nine attributes, specifically economic growth, energy consumption, research and development, financial development, foreign direct investment, trade openness, industrialization, and urbanisation, as input variables. The aforementioned elements play a pivotal role in influencing the level of carbon emission intensity.

Wu et al. (2020) performed an analysis on the carbon dioxide emissions of the BRICS countries, namely Brazil, Russia, India, China, and South Africa. The researchers utilised a conformable fractional non-homogeneous grey model in their investigation. The solutions of the novel model have been derived utilising mathematical methodologies and grey theory. Additionally, the ant lion optimizer, a meta-heuristic algorithm, has been employed to explore the optimal fractional order. Machine learning predictive models for predicting particulate matter concentrations in atmospheric air on a Taiwan air quality monitoring dataset obtained from 2012 to 2017 have been investigated by Doreswamy et al. (Doreswamy et al. 2020).

The study conducted by Qiao et al. (2021) employed the Discrete Grey Forecasting Model (DGM) to predict carbon dioxide ($CO_2$) emissions in the member nations of the Asia-Pacific Economic Cooperation (APEC) for the period of 2020-2023. The model utilised data from 2014 to 2019, and its performance was evaluated based on the Mean Absolute Percentage Error (MAPE) metric. In a separate study, Rehman and colleagues (2021) conducted an investigation of the effects of carbon dioxide emissions on many aspects, including forest productivity, crop production, animal production, energy consumption, population increase, temperature, and rainfall in the context of Pakistan. In their study, Yan et al. (2021) endeavoured to construct a predictive model for Beijing air quality that encompasses several locations and sites. To achieve this, they employed deep learning network models that incorporated spatial and temporal clustering techniques. Furthermore, the researchers conducted a comparative analysis between these models and a neural network known as BPNN. Espinosa et al. (2021) introduced a novel approach that utilises accuracy and robustness as key criteria for evaluating and contrasting various pollutant prediction models and their respective attributes. This study examines various deep learning models, including DCNN1, GRU, and LSTM, as well as regression models, such as random forest regression, lasso regression, and support vector machines, using different window widths. Salam et al. (2021) introduced a novel model called LSTM-SDAE (CLS) loop, which utilises deep learning techniques to forecast particulate matter levels. The model also uncovers the relationship between particulate matter and meteorological parameters.

The primary concentration of carbon dioxide ($CO_2$) emissions connected to energy production is observed in metropolitan regions. The significance of undertaking quantitative research on the association between carbon dioxide ($CO_2$) emissions and economic growth at both the municipal and sub-municipal levels is of utmost relevance. The study conducted by Shi et al. (2022) investigated the extent to which $CO_2$ emissions were disconnected from economic growth in 16 regions of Beijing throughout the period of 2006 to 2017. The Tapio decoupling model was utilised for this analysis. Chong et al. (2022) provides an overview of the latest advancements in several growing energy industries, with a particular focus on the importance of achieving carbon neutrality and ensuring energy sustainability in the period following the Covid-19 pandemic.

The significance of precise forecasting of air pollution levels is underscored in the existing body of literature, as it serves as a crucial component within the early warning system. Nonetheless, this issue continues to present a formidable obstacle, mostly stemming from the scarcity of available data regarding the emission source, as well as the substantial level of uncertainty surrounding the intricate dynamic processes involved (Kim et al. 2021). Analysis and prediction of the emission of pollutants and greenhouse gases are of double necessity in making decisions and preventing environmental destruction.

Predictive models based on artificial intelligence (AI) and machine learning (ML) techniques have been extensively employed and suggested for the estimation of pollutant parameters, with a primary focus on predicting $PM_{2.5}$ levels. As previously said, this discipline necessitates the incorporation and suggestion of innovative hybrid predictive models that are founded on the principle of an SVR model, which has been enhanced by the utilisation of robust optimisation techniques. The main contribution of this research is the utilisation and enhancement of innovative hybrid support vector regression (SVR)-based models in the field of pollutant prediction, with a specific focus on $PM_{2.5}$. The choice was made to incorporate two well-established and highly regarded optimisation methodologies, specifically Particle Swarm Optimisation (PSO) and Grey Wolf Optimisation (GWO), into hybrid Support Vector Regression (SVR) models. This paper presents a novel methodology that involves the utilisation of two support vector regression (SVR) models, specifically Particle Swarm Optimization-SVR (PSO-SVR) and Grey Wolf Optimization-SVR (GWO-SVR), in order to forecast $PM_{2.5}$ levels. The utilisation of the Particle Swarm Optimisation (PSO) and Grey Wolf Optimisation (GWO) algorithms is implemented in order to optimise the hyperparameters 'C' and 'gamma' of the Support Vector Regression (SVR) model. The goal is to enhance the predictive capabilities of the model and achieve improved performance. The work demonstrates innovation through the introduction and implementation of hybridization techniques in SVR models for $PM_{2.5}$ forecasting.

The paper is structured in the following manner. Section 2 encompasses the materials and methods employed in this investigation, encompassing the study area and data description, the models utilised, and the performance measures employed. Section 3 encompasses the results and discussion. Section 4 entails the conclusions.

2. **Materials and methods**

This section describes the implemented models and discusses the SVR-based optimization techniques and validation criteria.

2.1. Study area and data description

The present work utilised the dataset obtained from the UCI website in order to validate our model. In January 2013, Beijing implemented an air pollution monitoring network as an integral component of its nationwide monitoring network. In Beijing, there exists a total of 36 sites dedicated to the monitoring of air quality. Among these sites, 35 are operated by the Beijing Municipal Environmental Monitoring Centre (BMEMC), while the remaining site is managed by the United States embassy located in Beijing (Zhang et al. 2017). The dataset under consideration encompasses six prominent air contaminants and six associated meteorological variables observed at multiple locations throughout Beijing. The dataset comprises hourly measurements of air pollutants collected from 12 stations dedicated to monitoring air quality. The air quality data was acquired from the Beijing Municipal Environmental Monitoring Centre. The meteorological data obtained at each air quality site aligns with the meteorological station located nearby, which is under the operation of the China Meteorological Administration. The period of time being examined extends from March 1, 2013, to February 28, 2017 (UCI Machine Learning Repository: Beijing Multi-Site Air-Quality Data Data Set, n.d.). In this study, the data from Aotizhongxin for the years 2013 and 2014 have been selected for the purpose of model validation.

The American Environmental Protection Agency (EPA) has selected six primary pollutants as standard pollutants to measure the level of air pollution. Additionally, the data has been classified into two distinct categories: primary and secondary. Primary pollutants are compounds that are directly emitted into the ambient air from sources. They include $CO$, $NO_2$, $SO_2$, PM, and PB pollutants, except for the latter one found in our dataset. Secondary pollutants refer to the things that arise from reactions in the earth's atmosphere, and $O_3$ can be mentioned in this category. Meteorological conditions have a significant effect on air pollution. The issue of air pollution can be analysed in terms of meteorological factors, which can be classified into two main categories: primary and secondary. The primary characteristics encompass wind direction (WD), wind speed, temperature, while the secondary parameters encompass precipitation, humidity, radiation, and visibility. The aforementioned metrics exhibit a substantial correlation with latitude, seasonality, and topographic characteristics. The degree of pollution is influenced by weather conditions, and conversely, air pollution has an impact on weather conditions. As an illustration, the presence of air pollution has the potential to diminish visibility, intensify the occurrence and length of dense fogs, and diminish the amount of solar energy reaching the Earth's surface. The levels of rainfall and relative humidity in urban areas have the potential to both rise and fall.

2.2. Implemented models

Different algorithms are introduced and used to build a prediction model at this stage. These algorithms are implemented in Python software, and the accuracy of each one is obtained in order to choose the best method.

2.2.1. Support vector regression

The support vector machine (SVM) was first proposed by Vapnik in 1999, based on the concepts of statistical learning theory. The method in question is widely acknowledged within the academic community as a form of guided learning. The kernel function possesses the capability to convert input vectors that are non-linear into a space with multiple features. A hyperspace is created within the feature space in order to effectively distinguish and separate the two distinct data kinds. The distinctive attributes of the decision level guarantee that Support Vector Machines (SVM) possess a strong capacity for generalization (Rui et al. 2019). Furthermore, support vector machines have been employed for the analysis of time series and regression tasks in many research and scenarios (Gao, Qi, and Yang 2024). The SVM algorithm can be categorised into two main variants: support vector classification machine and support vector regression machine. The former typically pertains to tasks involving the categorization of data and is employed for the purpose of making predictions. Raj (Raj 2020) mentioned that although support vector regression is rarely used, it has certain advantages, as listed below: (i) The algorithm exhibits robustness against outliers. (ii) The decision model may be readily modified and updated. (iii) The algorithm demonstrates strong generalisation capabilities, resulting in accurate predictions. (iv) The implementation of the algorithm is straightforward and uncomplicated.

The SVR function possesses the ability to demonstrate both linear and non-linear behaviour. The Support Vector Regression (SVR) model employs a series of linear functions, characterised by the equation f(x) = (w.x) + b, in order to generate predictions. The equation presented herein involves the utilisation of variables x, w, and b, which respectively denote the input vector, weight vector, and bias term. The incorporation of a loss function is a fundamental component of this methodology, as it functions to quantify the permissible degree of disparity between the predicted values and the actual values (Drucker et al., 1996). Hence, the following equations are utilised to minimise the optimisation problem.

$$Minimize: \frac{1}{2}\|W\|^2 + C \sum_{i=1}^{n}(\xi^* + \xi) \quad (1)$$

$$Subject\ to: \begin{cases} Y_i - (W.X_i + b) \leq \mathcal{E} + \xi \\ (W.X_i + b) - Y_i \leq \mathcal{E} + \xi^* \\ \xi^*, \xi \geq 0 \end{cases} \quad (2)$$

In the context of a loss function, the symbol $\mathcal{E}$ denotes the permissible error, while $\xi$ and $\xi^*$ represent the variables that approach their respective limits. Additionally, C denotes the penalty parameter. It is important to acknowledge that the efficacy of Support Vector Regression (SVR) is contingent upon the appropriate configuration of certain parameters, including C, $\mathcal{E}$, and the relevant kernel parameters (Paryani et al., 2021).

The optimisation issue mentioned above can be transformed into a quadratic dual optimisation problem by using the Lagrange coefficients $\alpha_i$ and $\alpha_i^*$. Upon successfully solving the dual optimisation problem, the resultant parameter vector w is acquired in equation (3). The support vector regression (SVR) function is derived as equation (4).

$$W^* = \sum_{i=1}^{n}(\alpha_i - \alpha_i^*)(X_i) \quad (3)$$

$$f(X, \alpha_i, \alpha_i^*) = \sum_{i=1}^{n}(\alpha_i - \alpha_i^*)K(X_i, X_j) + \boldsymbol{b} \quad (4)$$

The Lagrange coefficients, denoted as $\alpha_i$ and $\alpha_i^*$, are utilised in conjunction with the kernel function $K(X_i, X_j)$ to facilitate non-linear mapping. Various kernels are employed in the Support Vector Regression (SVR) model. According to Hamzeh et al. (Hamzeh et al. 2017), some common kernels are:

$$K(X_1, X_2) = X_1^T X_2 \qquad \text{Linear kernel} \quad (5)$$

$$K(X_1, X_2) = (X_1^T X_2 + \gamma)^d \qquad \gamma, d > 0 \qquad \text{Polynomial kernel} \quad (6)$$

$$K(X_1, X_2) = exp\,(-\gamma \|X_1 - X_2\|^2) \qquad \gamma > 0 \qquad \text{Radial Basis Function (RBF)} \quad (7)$$

$$K(X_1, X_2) = tanh\,(\gamma X_1^T X_2 + r) \qquad \gamma, r > 0 \qquad \text{Sigmoid kernel} \quad (8)$$

The kernel parameters are denoted by r, γ, and d. The performance, generalizability, and accuracy of SVR models are contingent upon the optimal selection of parameters such as γ, r, C, and d.

2.2.2. Particle swarm optimization

The Particle Swarm Optimisation (PSO) algorithm is considered to be a highly effective approach for addressing optimisation problems, particularly when compared to other evolutionary search methods that mimic the behaviour of fish schools and bird colonies (Kennedy and Eberhart, 1995). Consequently, the researchers endeavour to enhance the accuracy of pollutant prediction outcomes by integrating the aforementioned approach with the Support Vector Regression (SVR) technique. In Particle Swarm Optimisation (PSO), a collection of particles is metaphorically represented by a flock of birds, while a food source symbolises a functional objective. Once the pertinent details regarding the spatial separation between the avian assemblages and the sustenance origin have been conveyed, the precise whereabouts of the sustenance origin can be ascertained through the congregations of avian groups. This collaborative behaviour enables the entire group of avian organisms to effectively communicate and determine the most accurate details on the whereabouts of the nourishment site, ultimately resulting in their collective convergence towards the food source. By employing these procedures, it is possible to furnish the most prevalent source of sustenance (Li et al. 2021).

In the Particle Swarm Optimisation (PSO) algorithm, the initialization phase involves assigning numerical values to the particles. Each particle is then considered as a potential candidate solution to the specific problem, with an equal likelihood of being picked. Subsequently, it is vital to precisely ascertain two crucial attributes of every particle, specifically the revised velocity (V) and the unchanging position (X) (Poli et al. 2007). The fitness function assesses the fitness of individual particles, and the positions of the particles' masses are adjusted according to the fitness function's evaluation outcomes. Through successive iterations, the particle swarm algorithm converges towards the optimal position that maximises the predefined goal function as determined by the users (Li et al. 2021). The relevant parameters in Particle Swarm Optimisation (PSO) are updated in the following manner, allowing for the determination of the new position and velocity.

$$\begin{cases} V^{t+1} = \omega V^t + c_1 rand(A)(P_{best} - X^t) + c_2 rand(B)(G_{best} - X^t) \\ X^{t+1} = X^t + V^{t+1} \end{cases} \qquad (9)$$

Where t be the current iteration number, and w rand(A) and rand(B) represent random numbers selected from the interval (0, 1). Pbest and Gbest represent the best separate particle and whole particle positions. $c_1$ and $c_2$ remain constants that control particle acceleration (Zhou et al. 2013). The symbol ω is used to denote the inertia weight, which plays a crucial role in determining the equilibrium between global and local optimization (Shi and Eberhart 1998). In general, the value of ω decreases in each iteration. It can be determined as follows:

$$\omega^{t+1} = \omega^{max} - \frac{\omega^{max} - \omega^{min}}{Iteration_{max}} \tag{10}$$

$\omega^{max}$ denotes the greatest value of the inertia weight, whereas ω^min indicates the minimum value. Additionally, Iteration$_{max}$ signifies the maximum number of iterations or repetitions. The Particle Swarm Optimisation (PSO) algorithm is utilised within the Support Vector Regression (SVR) framework to optimise two significant meta-parameters, specifically C and γ. The achievement of global optimisation in particle swarm is ultimately realised through the iterative procedure of updating the velocity and position of all particles inside the swarm. The overall procedure of Particle Swarm Optimisation (PSO) can be elucidated by the visual representation provided in Figure 1 (Li et al. 2021).

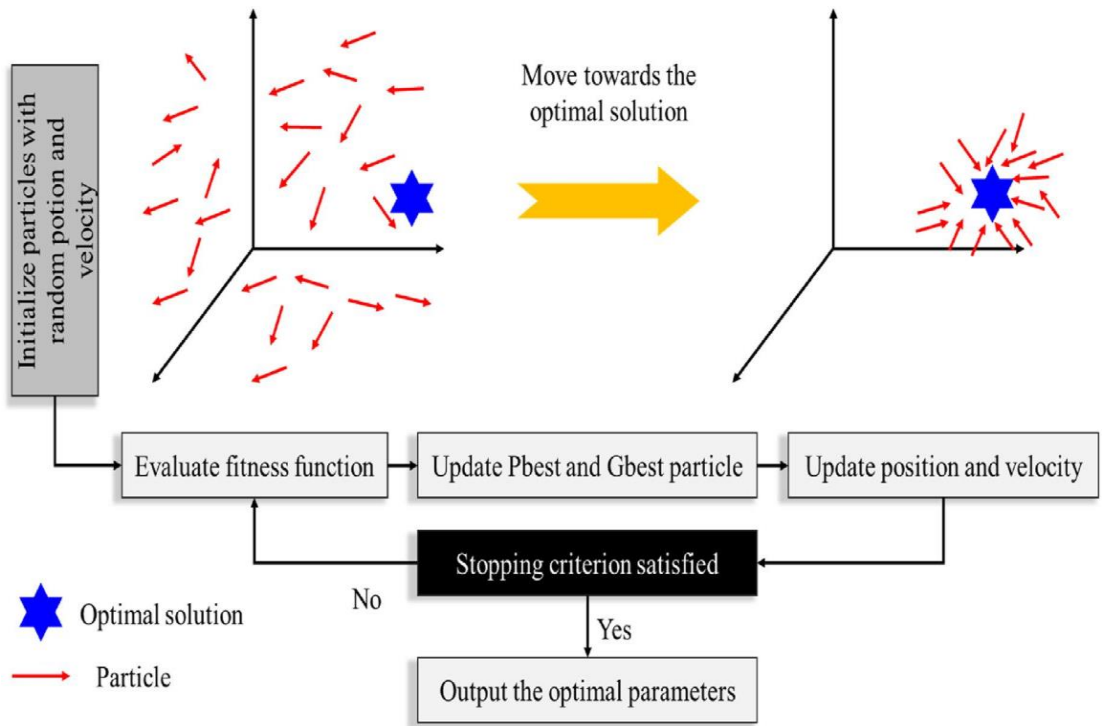

Fig. 1. A general process of PSO (after Li et al., 2021)

2.2.3. Grey wolf optimization

The GWO algorithm is a biologically inspired optimisation algorithm that emulates the social hierarchical leadership and hunting techniques observed in grey wolves (Mirjalili et al. 2014). The GWO algorithm yielded notable outcomes in comparison to other established algorithms. The outcomes pertaining to unimodal and multimodal functions provide evidence of the enhanced efficacy of the Grey Wolf Optimisation (GWO) algorithm. The outcomes of the integrated functions exhibit a significant tendency to avoid local optima, and the examination of Grey Wolf Optimisation (GWO) convergence verifies the convergence of this technique. The outcomes of engineering design challenges further demonstrate that the Grey Wolf Optimisation (GWO) algorithm exhibits exceptional performance when operating in unfamiliar and demanding search domains.

The Generalised World Optimisation (GWO) algorithm possesses various advantageous characteristics when applied to non-linear and multivariate functions. These include simplicity, flexibility, and the ability to avoid local optima, as highlighted by (Song et al. 2015). Grey wolves have a preference for residing in social groups consisting of 5 to 12 members (Emary, Zawbaa, and Hassanien 2016). Every individual wolf within the pack is assigned distinct responsibilities that the leader of the pack determines. Consequently, these entities are categorised into four distinct classifications, namely α, β, δ, and ω. The GWO algorithm is founded upon a hierarchical structure. Once a random solution (population) has been generated, the values of α, β, and δ are decided based on the most appropriate solutions. The determination of the value of ω during the remaining solutions is based on the equations provided by(Balogun et al. 2021):

$$\vec{X}(t+1) = \frac{\vec{X_1} + \vec{X_2} + \vec{X_3}}{3} \qquad (11)$$

$$\begin{cases} \vec{X_1} = \vec{X_\alpha} - \vec{A_1} \times (D_\alpha) \\ \vec{X_2} = \vec{X_\beta} - \vec{A_2} \times (D_\beta), \vec{A} = 2 \times \vec{a} \times \vec{r_1} - \vec{a}, \vec{D} = |\vec{C} \times \vec{X_p}(t) - \vec{X}(t)|, \vec{X}(t+1) = |\vec{X_p}(t) - \vec{A} \times \vec{D}| \qquad (12) \\ \vec{X_3} = \vec{X_\delta} - \vec{A_3} \times (D_\delta) \end{cases}$$

$$\begin{cases} \vec{D_\alpha} = |\vec{C_1} \times \vec{X_\alpha} - \vec{X}| \\ \vec{D_\beta} = |\vec{C_2} \times \vec{X_\beta} - \vec{X}|, \vec{C} = 2 \times \vec{r_2} \qquad (13) \\ \vec{D_\delta} = |\vec{C_3} \times \vec{X_\delta} - \vec{X}| \end{cases}$$

$\vec{X}$ and t represent the position of the wolf and the number of iterations. $\vec{X}_p$ is the position vector of the prey $\vec{A}$ and $\vec{C}$ represent the coefficient vectors and components which decrease linearly between 0 and 2 in each iteration (Tu, Chen, and Liu 2019). $r_1$ and $r_2$ are random vectors generated for the range [0,1] (Gupta and Deep 2019) (Figure 2). Hunting is also completed when a takes values between -1 and 1 when an attack occurs (Balogun et al. 2021).

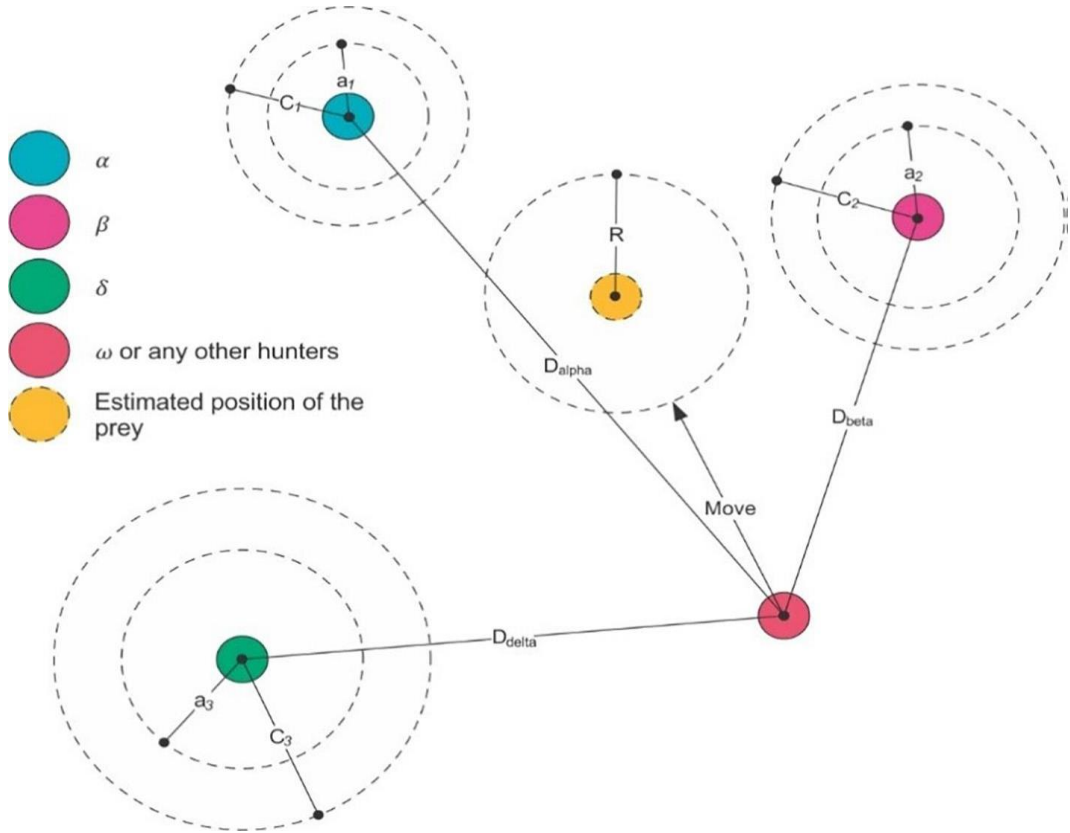

Fig. 2. Updating the position in the GWO algorithm (after Mirjalili et al., 2014).

2.3 SVR-based optimization techniques

The present study utilises Particle Swarm Optimisation (PSO) and Grey Wolf Optimisation (GWO) algorithms for the purpose of hyperparameter optimisation in a prediction model that is based on Support Vector Regression (SVR). After conducting numerous experiments, it has been observed that during each optimisation process, the computational time of the model tends to increase as the population sizes become larger with an increase in the number of iterations. The stability of fitness values is expected to be higher in populations with small sizes. In this paper, the decision was made to utilise a population size of 150 in the optimisation model for the purpose of generating models.

The hybrid model incorporating Support Vector Regression (SVR) utilises the Particle Swarm Optimisation (PSO) and Grey Wolf Optimisation (GWO) techniques to effectively optimise the hyperparameters 'C' and 'gamma' associated with the SVR model. Typically, the parameters are assigned values within the range of (0.01, 100). The fundamental procedure for optimising support vector regression (SVR) parameters utilising particle swarm optimisation (PSO) and grey wolf optimisation (GWO) approaches is outlined as follows:

(1) Data preparation: The dataset is partitioned into training and testing sets using a suitable 80% and 20% ratio.

(2) Initialization parameters: The parameters for Particle Swarm Optimisation (PSO) and Grey Wolf Optimisation (GWO) are established as shown in Table 1.

(3) Fitness evaluation: The fitness function will be computed, and its fitness will be assessed prior to optimising the value of the target parameter.

(4) Update parameters: Based on the outcomes seen in each iteration, it is necessary to modify the optimisation criteria that the hyperparameters should satisfy.

(5) Stop condition checking: The optimal parameters are achieved when the optimisation termination criterion is met.

**Table 1:** Parameter configurations of meta-heuristic algorithm

| Meta-heuristic algorithm | Parameter | Value |
|---|---|---|
| GWO | A | Decreasing linearly from 2 to 0 |
|  | Ub | 50 |
|  | Lb | 0.01 |
| PSO | $c_1$ | 1 |
|  | $c_2$ | 2 |
|  | W | 0.5 |

2.3.1. PSO-SVR model

One limitation of Support Vector Regression (SVR) is that it imposes certain constraints that may restrict its applicability in academic and industrial settings. The researcher must define some free parameters, namely the SVR hyperparameters and SVR kernel parameters. The efficacy of SVR regression models is contingent upon the appropriate configuration of its parameters. Consequently, practitioners face the primary challenge of determining the optimal parameter values to achieve favourable generalisation performance when applying SVR to a specific training dataset. The pseudocode for the PSO-SVR algorithm is presented in Table 2.

**Table 2:** Mechanism of PSO-SVR

```
P = particle Initialization ();
For i=1 to itr_max
    For each particle p in P do
        fp = f(p);
        If fp is better than f(p_Best);
            p_Best = p;
        end
    end
    g_Best = best p in P
    Determine the non-dominated objective
values
    Update the no. of non-dominated
solutions in the archive
    For each particle p in P do
        v = v + c_1 *rand*(p_Best – p) + C_2
*rand*(g_Best-P);
        p = p+v;
    end
end
Print the best solution.
```

The PSO algorithm with the training samples determines the optimal parameter combination of SVR. The testing samples confirm the effectiveness of the PSO-SVR regressor. The establishment of the PSO-SVR model is described in Figure 3.

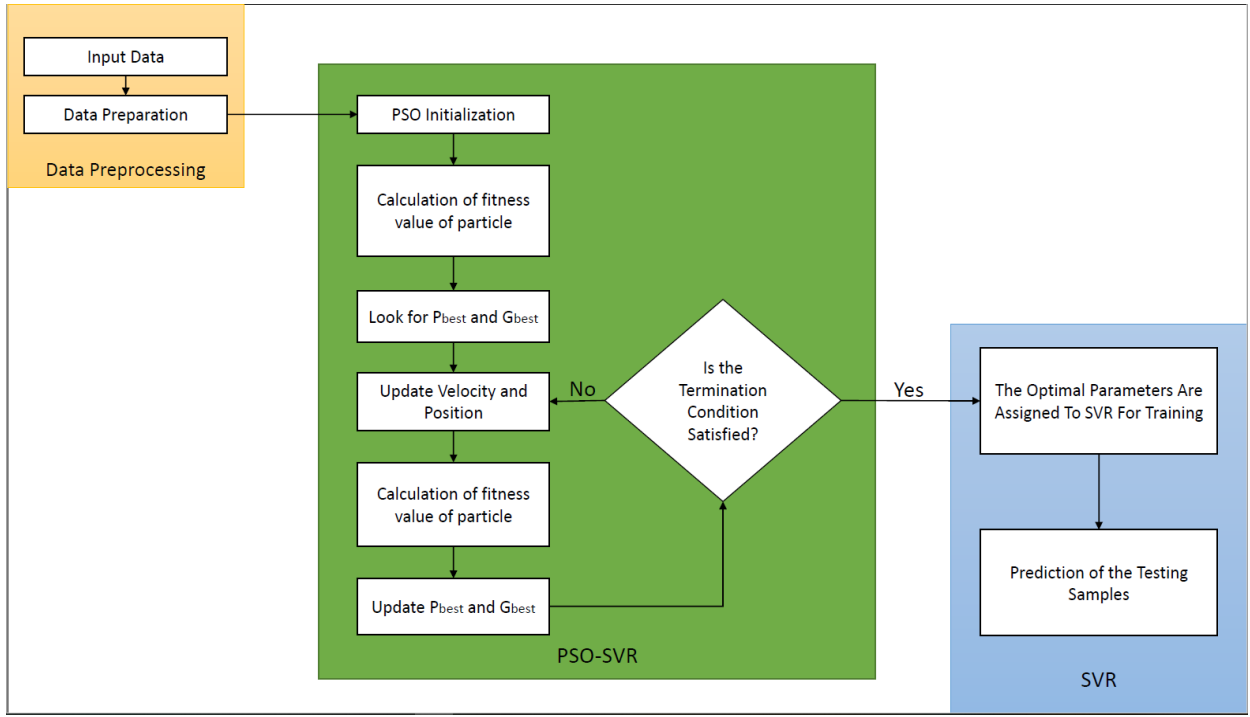

Fig. 3. Schematic of PSO-SVR

2.3.2. GWO-SVR model

The major focus of prior research has been on utilising the Genetic Algorithm (GA) and Particle Swarm Optimisation (PSO) methods to optimise parameters inside the Support Vector Regression (SVR) model. However, it should be noted that these optimisation strategies often demonstrate slow convergence rates, complex parameter settings, or a tendency to get stuck in local optima. Therefore, the current work utilises the Grey Wolf Optimisation (GWO) algorithm to optimise the parameters of the Support Vector Regression (SVR) model. The GWO approach demonstrates a decreased quantity of parameters and contains a significant level of global search capability. The execution of this approach is uncomplicated and efficiently governs the local search range of the algorithm, so attaining a harmonious equilibrium between its global search capacity and local search capacity. The schematic representation of the GWO-SVR model is depicted in Figure 4.

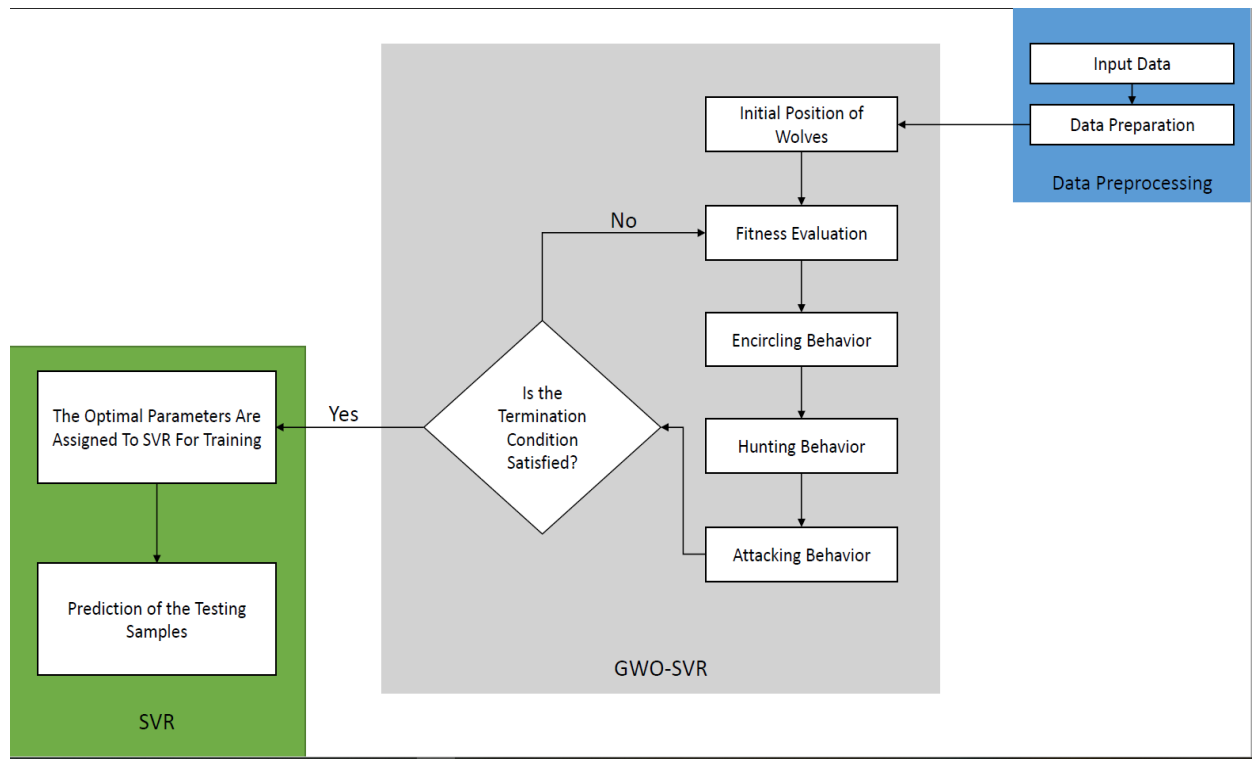

Fig. 4. Schematic of GWO-SVR

According to previous research, it has been demonstrated that larger sample sizes are associated with improved model performance and greater convergence when utilising the RBF and Sigmoid kernel functions, as opposed to the polynomial kernel function. In situations when the sample size is constrained and the number of features significantly surpasses the number of samples, it is plausible for the linear kernel function to provide performance that is on par with the radial basis function (RBF) kernel. Therefore, it can be obtained. Regardless of the presence of various characteristics, such as small features, multiple samples, or small sample sizes, it is evident that the RBF kernel function has excellent performance in modelling and boasts a strong capability for non-linear mapping. In this study, the RBF kernel function has been chosen as the kernel function for the training prediction model of Support Vector Regression (SVR). The pseudocode of the GWO-SVR method is displayed in Table 3.

**Table 3:** Mechanism of GWO-SVR

```
Input: the number of grey wolf population N, the maximum number of iterations T, the number of parameters solved p,
the optimal value range of penalty parameter c, RBF kernel function parameter g, and the data set D.
Output: the optimal parameters Best_c and Best_g, the predicted value, and the error value of SVR.
Initialize the grey wolf population X_i (i = 1, 2, ..., n)
Initialize the fittest solution α, search coefficient A and C
call the SVR
Calculate the fitness of each search agent
X_α = the best search agent
X_β = the second search agent
X_δ = the third search agent
while (t < T)
    for each search agent
        Update the position of the current search agent by equation (11)
    end for
    Update α, A, and C
    call the SVR
    Calculate the fitness of all search agents
    Update X_α, X_β, and X_δ
    t = t +1
end while
```

2.4 Performance metrics

Three mathematical evaluation metrics are adopted for the validation of the proposed models. In general, the optimal prediction performance is indicated by RMSE and MAE values of zero, whereas R2 values of 100. Various optimisation algorithms yield distinct prediction outcomes, making it possible to employ these values in order to ascertain the most effective optimisation technique. In the present study, a comprehensive evaluation methodology is utilised to analyse the overall performance of the three algorithms under consideration.

$$R^2 = 1 - \frac{\sum_{i=1}^{M}(y_i - y'_i)^2}{\sum_{i=1}^{M}(y_i - y''_i)^2} \tag{14}$$

$$RMSE = \sqrt{\frac{\sum_{i=1}^{M}(y_i - y'_i)^2}{M}} \tag{15}$$

$$MAE = \frac{\sum_{i=1}^{M}|y_i - y'_i|}{M} \tag{16}$$

Where $y_i$, $y'_i$, and $y''_i$ represent the original, predicted, and mean values of PM$_{2.5}$, and M represents the total amount of data.

## 3. Results and discussion

In order to investigate more effective prediction techniques for $PM_{2.5}$, an initial approach involved the independent use of Support Vector Regression (SVR) for prediction. Subsequently, two optimisation algorithms, namely Particle Swarm Optimisation (PSO) and Grey Wolf Optimisation (GWO), were integrated with SVR to enhance the prediction process. The construction of these hybrid intelligent models, based on Support Vector Regression (SVR), was carried out utilising the training data. The optimisation procedure described above yielded distinct hyperparameter combinations and varied model prediction performances.

The relationship between the estimated and observed values of $PM_{2.5}$ for the years 2013 and 2014 is depicted in Figures 5 and 6, correspondingly. The findings indicate that the intelligent models yield outstanding results, with the sample points closely aligned with the ideal fitting line representing the relationship between actual and forecast $PM_{2.5}$ values. The performance index findings (Root Mean Square Error, R-squared, and Mean Absolute Error) and complete ranking results of the models (Support Vector Regression, Grey Wolf Optimizer-Support Vector Regression, and Particle Swarm Optimization-Support Vector Regression) in their ability to forecast $PM_{2.5}$ are summarised in Tables 1 and 2. The findings of the models indicate that there are significant differences in the overall scores between PSO-SVR and GWO-SVR. The SVR hybrid models have better accuracy and robustness in predicting $PM_{2.5}$ compared to SVR on its own. The enhanced precision and resilience observed in hybrid Support Vector Regression (SVR) models such as PSO-SVR and GWO-SVR can be ascribed to the incorporation of optimization methods, specifically Particle Swarm Optimization (PSO) and Grey Wolf Optimizer (GWO), within the SVR framework. The utilization of optimization approaches improves the model's capacity to finely adjust parameters and effectively capture intricate patterns present in the $PM_{2.5}$ data. The incorporation of hybrid models is expected to enhance the alignment with the fundamental data distribution, hence leading to improved precision in forecasting.

Moreover, the significance placed on the "comprehensive ranking results" suggests that the assessment is not limited to a solitary performance measure, but rather incorporates a whole comprehension of model behavior. The utilization of a multi-metric technique offers a more equitable assessment of the models' capacities, guaranteeing that the chosen model not only demonstrates exceptional performance in one particular area but also exhibits satisfactory results across a range of evaluation criteria.

In summary, the comprehensive analysis reinforces the claim that hybrid support vector regression (SVR) models, specifically PSO-SVR and GWO-SVR, exhibit superior performance compared to conventional SVR in the prediction of $PM_{2.5}$ concentrations. The strong convergence of data points in close proximity to the optimal regression line depicted in the correlation plots, along with the consistent and reliable performance across all evaluation criteria, collectively provide substantial evidence to substantiate the assertion of the models' exceptional quality. The present study highlights the potential advantages of incorporating optimization techniques into regression models for the purpose of predicting environmental contaminant levels.

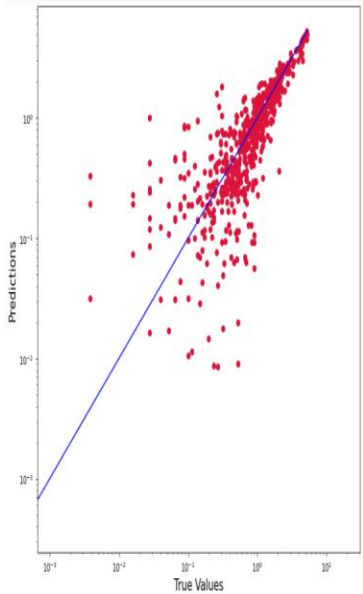 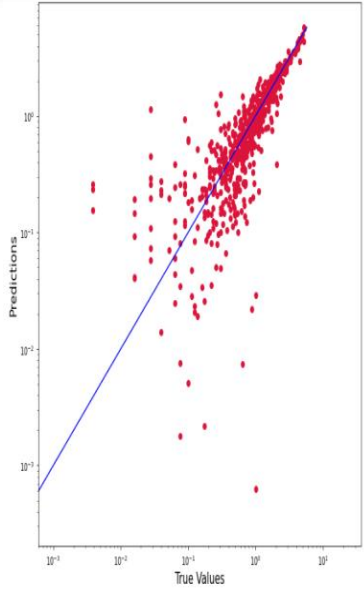 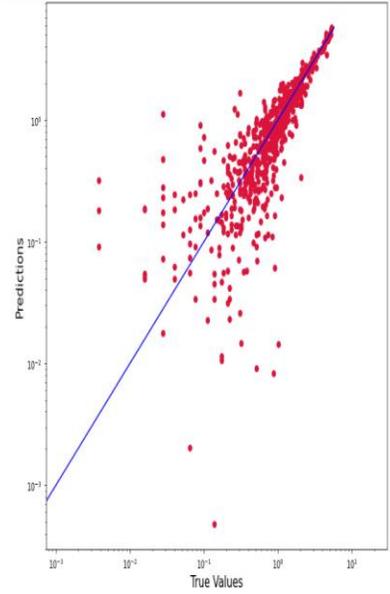

Fig. 5-a. SVR (2013)　　　Fig. 5-b. PSO-SVR (2013)　　　Fig. 5-c. GWO-SVR (2013)

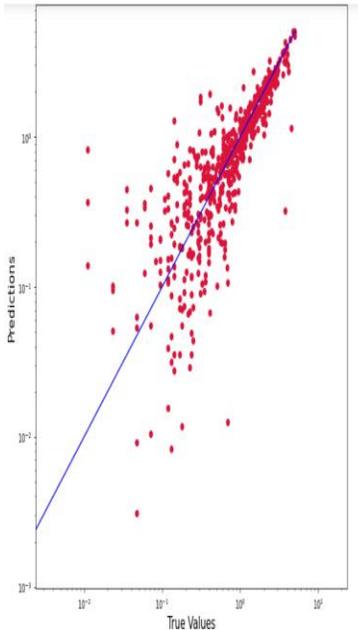 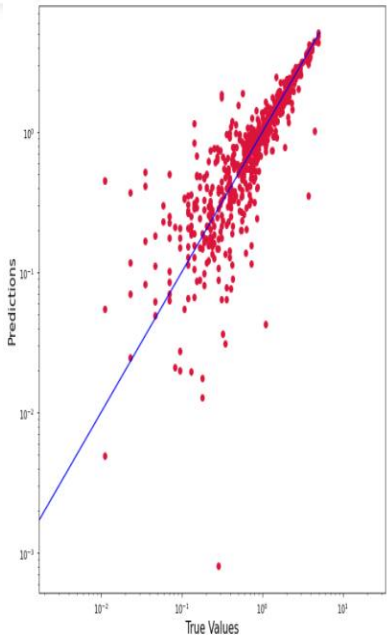 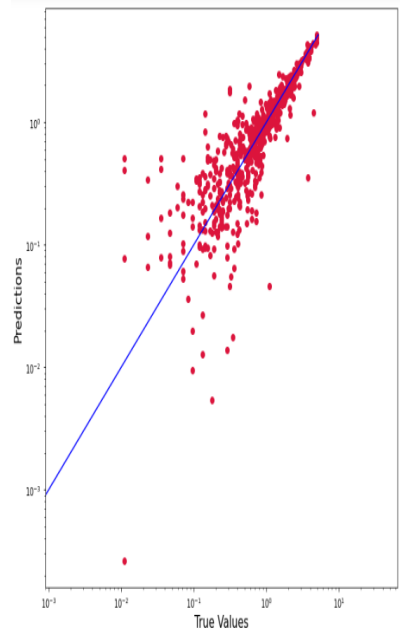

Fig. 6-a. SVR (2014)　　　Fig. 6-b. PSO-SVR (2014)　　　Fig. 6-c. GWO-SVR (2014)

**Table 4:** Performance of each model for 2013 & 2014 records.

| Metric<br>Model | 2013 | | | 2014 | | |
|---|---|---|---|---|---|---|
| | $R^2$ | RMSE | MAE | $R^2$ | RMSE | MAE |
| SVR | 0.9312 | 0.2646 | 0.176 | 0.9257 | 0.2662 | 0.1559 |
| PSO-SVR | 0.9397 | 0.2477 | 0.165 | 0.9401 | 0.2390 | 0.1368 |
| GWO-SVR | 0.9389 | 0.2493 | 0.166 | 0.9408 | 0.2376 | 0.1373 |

From the perspective of $R^2$, RMSE, and MAE, as shown in Table 4, the prediction performance of the hybrid models in 2014 is slightly better than in 2013 because of missing values in 2013 and 2014 (Table 5). It should be mentioned that the missing values are replaced with the mode values in each year's records. The hybrid SVR models can significantly increase the performance capacity of a pre-developed SVR model in estimating $PM_{2.5}$. For instance, developing SVR-based models can reduce the RMSE value 2013 from about 0.2646 to 0.2477. The finding of hybrid Support Vector Regression (SVR) models regularly demonstrate superior performance compared to the pre-developed SVR model is a significant observation. The decrease in root mean square error (RMSE) from around 0.2646 to 0.2477 in the year 2013 serves as a notable demonstration of the significant influence of the hybrid strategy on enhancing the accuracy of predictions. The aforementioned decrease in error suggests that the hybrid Support Vector Regression (SVR) models possess the ability to better approximate the true $PM_{2.5}$ concentrations, hence enhancing the reliability of the forecasts. The integration of Support Vector Regression (SVR) with meta-heuristic optimization methods like Particle Swarm Optimization (PSO) and Grey Wolf Optimization (GWO) enables the models to optimize their parameters in order to better align with the unique attributes of the data, resulting in enhanced performance.

Moreover, it is crucial to emphasize the practical ramifications of this enhancement. The enhancement of predictive accuracy for $PM_{2.5}$ concentrations holds significant significance in the realms of public health and urban planning. By minimizing the occurrence of prediction errors, decision-makers are able to enhance the quality of their choices pertaining to activities that could potentially be influenced by air quality, such as outdoor gatherings or building projects. The potential of hybrid models to improve forecast accuracy has the capacity to alleviate the adverse health impacts of air pollution and contribute to overall well-being.

In brief, the enhancements in predictive accuracy witnessed from 2013 to 2014 can be attributed to the effective management of missing data and the incorporation of optimization methods into the hybrid Support Vector Regression (SVR) models. The decrease in root mean square error (RMSE) demonstrates the practical importance of these improvements. The research holds significant importance in resolving the issues faced by air pollution in urban settings, as it has the potential to yield more precise predictions. This, in turn, has far-reaching consequences for public health and urban development.

**Table 5:** Missing values of 2013 & 2014 records.

|  | 2013 | | 2014 | |
|---|---|---|---|---|
| **Parameter** | Missing-Number | Missing-Percent | Missing-Number | Missing-Percent |
| **$PM_{2.5}$** | 37 | 0.004223 | 505 | 0.057642 |
| **$PM_{10}$** | 30 | 0.003424 | 487 | 0.055587 |
| **$NO_2$** | 101 | 0.011528 | 614 | 0.070083 |
| **$SO_2$** | 138 | 0.015752 | 573 | 0.065403 |
| **CO** | 918 | 0.104783 | 585 | 0.066773 |
| **$O_3$** | 617 | 0.070426 | 597 | 0.068143 |
| **WD** | 1 | 0.000114 | 2 | 0.000228 |

Ultimately, the most successful SVR and SVR-based models are selected and subjected to a thorough comparison. Based on the aforementioned discussions, it is evident that PSO-SVR and GWO-SVR achieve the highest scores. Furthermore, PSO-SVR demonstrates superior performance across three metrics, namely RMSE, $R^2$, and MAE, in the year 2013. Additionally, PSO-SVR outperforms in one metric, specifically MAE, in the year 2014. Although GWO-SVR demonstrates superior performance in only two criteria, namely $R^2$ and RMSE, in the year 2014. The PSO-SVR and GWO-SVR models exhibit greater performance in the testing set, hence demonstrating their enhanced generalisation and resilience capabilities. In the context of AI-based models, the potential benefits that may arise from small advantages can be significantly amplified when applied to extensive datasets. Hence, based on the findings of this study, it can be concluded that PSO-SVR and GWO-SVR exhibit superior performance as the most effective approaches for PM2.5 prediction.

The examination of PSO-SVR and GWO-SVR models in greater detail reveals that the disparities in performance metrics across multiple years underscore the intricate characteristics of air quality forecasting. The resilience and promise for dependable forecasts of PSO-SVR are highlighted by its persistent superiority in three out of four metrics for the year 2013, as well as in one indicator for the year 2014. The observed pattern indicates that the PSO optimization process effectively captures the intricate associations between predictor variables and $PM_{2.5}$ concentrations. This is achieved through the meticulous adjustment of meta-parameters in the SVR model, enabling consistent performance throughout multiple years, despite probable variations in pollution patterns.

However, it is worth noting that although GWO-SVR only exhibits superior performance compared to PSO-SVR in two specific criteria for the year 2014, its results still hold potential significance and can provide significant insights. The observation that Grey Wolf Optimization has proficiency in specific measures suggests that it may possess a heightened ability to explore particular aspects of the model's parameter space, resulting in enhanced predictive capabilities within specific contexts. This suggests that the selection of an optimization procedure may not have universal applicability, but rather relies on the individual attributes of the data and the problem being addressed.

The assertion on the advantageous nature of AI-based models in relation to the utilization of extensive datasets holds considerable importance. As the size of the dataset increases, the nuances that play a role in the performance of the model become increasingly apparent. The persistent superior performance of both PSO-SVR and GWO-SVR models, particularly when subjected to stringent testing circumstances, highlights their potential for scalability and adaptability to bigger and more heterogeneous datasets. The resilience of these models can be ascribed to their capacity to adeptly assimilate patterns and variations present in the data, a critical factor for making precise long-term forecasts of air pollution levels.

In summary, it can be observed that both PSO-SVR and GWO-SVR have exceptional efficacy in the domain of air quality prediction. However, their subtle distinctions underscore the intricate nature of the underlying problem. The PSO-SVR model exhibits nuanced benefits across several parameters and over multiple years, indicating its potential for wider use and generalizability. Nevertheless, it is worth noting that the GWO-SVR model has distinct advantages in particular measures, underscoring the capacity of optimization algorithms to generate tailored solutions that align with the characteristics of the given dataset. The aforementioned discourse highlights the significance of meticulous selection and customization of optimization strategies in order to get optimal performance for air quality prediction models. This contributes to the enhancement of dependability and precision in forecasting $PM_{2.5}$ concentrations in urban settings.

## 6. Challenges and Limitations

The hybrid models presented for $PM_{2.5}$ prediction utilizing Support Vector Regression (SVR) and meta-heuristic algorithms exhibit notable breakthroughs. However, it is important to acknowledge and address various obstacles and limits associated with these models.

One of the primary obstacles is in the dependence on data that is both of superior quality and encompasses a wide range of information. The accuracy of air pollutant projections is greatly influenced by the availability of a comprehensive dataset that includes a range of parameters that impact pollution levels, including meteorological conditions, traffic patterns, and industrial activity. The absence or incorrectness of data has the potential to result in skewed model outputs and degraded predictive capabilities. Furthermore, it is important to note that the historical data employed for the purpose of training and evaluating the models may not comprehensively encompass the progressive dynamics of air pollution sources and patterns. This limitation could potentially hinder the models' ability to adapt to the ever-changing urban landscapes.

The efficacy of hybrid models relies on the meticulous choice and calibration of optimization methods, namely Particle Swarm Optimization (PSO) and Grey Wolf Optimization (GWO), for the purpose of refining Support Vector Regression (SVR) parameters. Nevertheless, the efficacy of these algorithms may be influenced by variables such as the initial parameter values and convergence conditions. It is imperative to ensure that the selected optimization algorithms has robustness, suitability for the specific situation, and appropriate configuration. Inadequately constructed algorithms may lead to inefficient parameter adjustments, hence reducing the overall predictive performance of the models.

The effectiveness of hybrid models could potentially be impacted by the specific attributes inherent in the dataset utilized for training purposes. The attainment of a high level of accuracy within a specific dataset does not always imply that similar levels of performance will be achieved in different locations or years characterized by unique pollution profiles. These models may be susceptible to overfitting, a phenomenon in which they inadvertently incorporate irrelevant or abnormal patterns from the training data, resulting in limited ability to accurately predict outcomes for novel and unobserved data. The significant problem lies in designing models that possess adequate flexibility to accommodate diverse situations, while still retaining their robust prediction skills.

Hybrid models that integrate support vector regression (SVR) with optimization techniques exhibit a higher level of complexity compared to standalone models, hence posing challenges in terms of interpretation and comprehension. The complex mechanisms underlying these models may present challenges in effectively communicating their functionality to stakeholders, hence impeding their acceptance and practical integration into decision-making procedures. The utilization of transparent models is of utmost importance in establishing the confidence and trust of policymakers, urban planners, and companies, as it enables them to comprehend the mechanisms behind forecast generation. Achieving a harmonious equilibrium between the intricacy of a model and its interpretability is a nuanced undertaking that necessitates meticulous deliberation.

In brief, although the hybrid models being suggested present notable progress in the prediction of air pollutants, they encounter difficulties pertaining to the quality of data, sensitivity of optimization algorithms, generalization, and the complexity of the models. It is important to acknowledge and overcome these constraints in order to guarantee the practical applicability and dependability of the models in various urban settings.

## 7. Commercial Implications of this study

The results of this study hold considerable commercial ramifications for the fields of urban planning and public health. The issue of air pollution is becoming increasingly worrisome in highly populated urban areas, and the precise forecasting of pollutants such as $PM_{2.5}$ is of utmost importance in order to effectively address and minimize its detrimental impacts. The hybrid models that have been proposed, which combine Support Vector Regression (SVR) with Particle Swarm Optimization (PSO) and Grey Wolf Optimization (GWO), present a new and innovative method for overcoming the limitations observed in earlier forecasting models. The implementation of these optimization techniques has the potential to significantly transform the approach to air quality management in urban areas and industrial sectors.

Accurate air pollutant forecasting holds significant importance from an urban planning standpoint as it enables informed decision-making about traffic management, industrial operations, and measures aimed at safeguarding public health. The findings of this study can be utilized by municipal authorities and urban planners to take proactive measures in mitigating pollution surges. These measures may include the implementation of traffic rules, optimization of industrial activities, and timely dissemination of health advisories to the public. The availability of accurate hourly concentration forecasts enables urban areas to execute specific interventions aimed at mitigating pollution levels, hence fostering a healthier and more sustainable urban environment.

This discovery also holds potential benefits for industries, particularly those operating in sectors that are associated with the emission of air pollutants. The prediction models outlined in this research can assist many businesses in anticipating periods characterized by elevated levels of pollution, enabling them to effectively modify their production schedules or implement appropriate emissions control systems. The adoption of a proactive approach can yield benefits for industries, encompassing both compliance with environmental rules and the improvement of their public image through the demonstration of a steadfast commitment to lowering their ecological footprint. Moreover, companies that specialize in environmental monitoring and pollutant control technologies can utilize the knowledge acquired from this research to create customized solutions that incorporate real-time pollutant data and predictive modeling. This will enable them to provide municipalities and industries with more efficient tools for managing air quality.

Moreover, the suggested hybrid models exhibit a high level of resilience and application, rendering them viable options for inclusion into commercial platforms dedicated to monitoring and forecasting air quality. Collaborative efforts between companies specializing in environmental monitoring technologies, data analytics, and software development can be undertaken to jointly create user-friendly applications that offer real-time pollution forecasts to both individuals and corporations. These applications have the potential to empower users in making well-informed decisions regarding outdoor activities, adapting commuting routes, and implementing preventive health measures in times of heightened pollution levels. These platforms have the potential to generate revenue through several means, such as implementing subscription models, establishing collaborations with municipal governments, or entering into licensing deals with enterprises aiming to improve their environmental sustainability.

In summary, the present study's novel methodology for predicting air pollutants, in conjunction with the improved precision attained by hybrid support vector regression (SVR) models and optimization methodologies, presents significant commercial prospects. The utilization of these technologies not only grants urban planners, city officials, and industries the capability to actively oversee air quality, but also fosters potential for enterprises to innovate and provide sophisticated environmental monitoring and prediction solutions to a diverse array of stakeholders.

## 8. Conclusions and Fututre Directions

In the present study, the utilisation of hybridised Support Vector Regression (SVR) models was employed to predict $PM_{2.5}$ values. The research employed two established optimisation methodologies, specifically Particle Swarm Optimisation (PSO) and Grey Wolf Optimisation (GWO), which have been previously examined by other scholars. The integration of these methodologies was afterwards accomplished through the utilisation of Support Vector Regression (SVR). Following that, researchers built hybrid models that included Particle Swarm Optimization-Support Vector Regression (PSO-SVR) and Grey Wolf Optimization-Support Vector Regression (GWO-SVR) in order to improve their predictive powers. This study investigated the fundamental factors influencing the Particle Swarm Optimisation (PSO) and Grey Wolf Optimisation (GWO) algorithms, resulting in the discovery of the parameters that exerted the most substantial influence. The data indicate that the hybrid models had the highest level of accuracy in predicting performance. The assessment of the SVR-based models was performed using evaluation measures including Root Mean Square Error (RMSE), Coefficient of Determination ($R^2$), and Mean Absolute Error (MAE). Following an extensive assessment of multiple established and novel models, it was ascertained that the PSO-SVR and GWO-SVR models demonstrated outstanding performance. The aforementioned models demonstrated $R^2$ values of 0.9401 and 0.9408, RMSE values of 0.2390 and 0.2376, and MAE values of 0.1373 and 0.1368, correspondingly. Hence, the SVR-based models presented in this research can be applied in other endeavours involving the prediction of $PM_{2.5}$. It is important to acknowledge that additional data and analysis are required in order to effectively anticipate $PM_{2.5}$ levels in various extreme scenarios. The use of the hybrid model put forth in this scholarly article is advised solely in circumstances that closely align with the conditions outlined and within a rational scope of database information.

To enhance the prediction capability of the model, it is recommended to employ a more comprehensive experimental database in the future, encompassing a larger number of samples and incorporating more features. Furthermore, it should be noted that strategies based on artificial intelligence have limitations in their ability to fully replace traditional methods that have proven to be effective. In the field of engineering, the future trajectory of AI technology is oriented towards the advancement of composite systems, specifically focusing on the creation of decision support tools. It is important to note that the clever procedures employed in this study are specifically advised for application in comparable circumstances. One primary constraint associated with these methodologies in this particular domain pertains to the utilisation of site-specific data for the formulation of artificial intelligence models. A promising avenue for future investigation involves the integration of the suggested hybrid support vector regression (SVR) models, which have been refined using particle swarm optimization (PSO) and grey wolf optimization (GWO), into real-time air quality monitoring systems. The proposed endeavor entails the creation of a system that consistently gathers data from many sources, including air quality monitors, meteorological stations, and traffic monitoring devices. The hybrid models have the potential to be utilized for the prediction of $PM_{2.5}$ concentrations in the forthcoming hours or days. The implementation of such a system has the potential to offer the public with air quality projections that are both timely and accurate. This would enable individuals to proactively adopt preventive measures and make well-informed choices regarding outdoor activities.

An additional area of research that shows potential is the improvement of the precision of the hybrid Support Vector Regression (SVR) models by integrating spatiotemporal elements. The levels of air pollution within a city might exhibit temporal fluctuations as well as spatial variations. The inclusion of spatial information, encompassing geographical characteristics, land utilization patterns, and data on transportation congestion, has the potential to enhance the model's ability to account for localized disparities in $PM_{2.5}$ concentrations. Furthermore, incorporating temporal patterns, including daily and weekly fluctuations, together with the impact of seasonal variations, has the potential to enhance the precision of the forecasts. This may entail employing sophisticated machine learning methodologies such as convolutional neural networks (CNNs) or recurrent neural networks (RNNs) for the purpose of processing spatiotemporal data.

In order to improve the resilience and capacity for generalization of the hybrid models proposed, future research endeavors may consider investigating the application of ensemble approaches. Ensemble models are a technique that leverages the predictions generated by numerous models in order to achieve a higher level of accuracy and reliability in the final outcome. Researchers have the potential to create a collection of diverse forecasting models, which may consist of the suggested Support Vector Regression (SVR) models, alongside other well-established methodologies such as neural networks, time series analytic approaches, and conventional statistical models. The utilization of an ensemble technique has the potential to reduce the vulnerability associated with over-reliance on a singular model, hence enhancing the stability of predictions. In addition, conducting experiments on the proposed models in other cities characterized by differing degrees of pollution and unique urban features could serve to substantiate their efficacy in diverse settings. By considering these prospective avenues, the scholarly article has the potential to enhance air quality forecasting models and their pragmatic application, ultimately resulting in enhanced public health outcomes and more effective air pollution control in urban regions.

**Conflicts of interest**

The authors declare that they have no known competing financial interests or personal relationships that could have appeared to influence the work reported in this paper.

**Data Availability**

The data used in this research is publicly available on the UCI Machine Learning Repository at the following address:

https://archive.ics.uci.edu/dataset/501/beijing+multi+site+air+quality+data